\documentclass{llncs}
\usepackage[utf8]{inputenc}
\usepackage{graphicx}
\usepackage{siunitx}
\usepackage{url}  
\usepackage{hyperref}
\usepackage{amsmath}
\usepackage{amssymb}
\usepackage{xspace}
\usepackage{cleveref}
\usepackage{listings}
\usepackage{float}
\usepackage{calc}
\usepackage{array}
\usepackage{color}
\usepackage{xcolor}
\newcolumntype{L}[1]{>{\raggedright\let\newline\\\arraybackslash\hspace{0pt}}m{#1}}
\newcolumntype{C}[1]{>{\centering\let\newline\\\arraybackslash\hspace{0pt}}m{#1}}
\newcolumntype{R}[1]{>{\raggedleft\let\newline\\\arraybackslash\hspace{0pt}}m{#1}}
\newcolumntype{P}[1]{>{\centering\arraybackslash}p{#1}}

\newcommand*{\eg}{e.\,g.\@\xspace}
\newcommand*{\ie}{i.\,e.\@\xspace}
\newcommand*{\cf}{cf.\@\xspace}

\makeatletter
\newcommand*{\etc}{%
    \@ifnextchar{.}%
        {etc}%
        {etc.\@\xspace}%
}

\definecolor{code_indent}{HTML}{CCCCCC}
\newcommand{\indentrule}{\color{code_indent}\vrule\hspace{2pt}}

\definecolor{red}{rgb}{0.6,0,0} 
\definecolor{blue}{rgb}{0,0,0.6}
\definecolor{green}{HTML}{44AA44}
\definecolor{cyan}{rgb}{0.0,0.6,0.6}
\definecolor{gray}{HTML}{333333}

\begin{document}
\title{Concept Drift Detection\\with Variable Interaction Networks}\footnotetext[1]{The final publication is available at \url{https://link.springer.com/chapter/10.1007/978-3-030-45093-9_36}}
%
%
\author{Jan Zenisek\inst{1,2},
Gabriel Kronberger\inst{1},
Josef Wolfartsberger\inst{1},\\
Norbert Wild\inst{1} \and
Michael Affenzeller\inst{1,2}}
\authorrunning{J. Zenisek et al.}
%
\institute{
  Center of Excellence for Smart Production\\
	University of Applied Sciences Upper Austria, Campus Hagenberg\\
	Softwarepark 11, 4232 Hagenberg, Austria\\
  \and
  Institute for Formal Models and Verification\\
  Johannes Kepler University Linz\\
  Altenberger Stra{\ss}e 69, 4040 Linz, Austria\\
  \medskip   
	\email{jan.zenisek@fh-hagenberg.at}
}
\maketitle              
%
%
%
\begin{abstract}
	The current development of today's production industry towards seamless sensor-based monitoring is paving the way for concepts such as Predictive Maintenance. By this means, the condition of plants and products in future production lines will be continuously analyzed with the objective to predict any kind of breakdown and trigger preventing actions proactively. Such ambitious predictions are commonly performed with support of machine learning algorithms. In this work, we utilize these algorithms to model complex systems, such as production plants, by focussing on their variable interactions. The core of this contribution is a sliding window based algorithm, designed to detect changes of the identified interactions, which might indicate beginning malfunctions in the context of a monitored production plant. Besides a detailed description of the algorithm, we present results from experiments with a synthetic dynamical system, simulating stable and drifting system behavior.
	
\end{abstract}
%
%
%
\section{Motivation}
\label{sec:motivation}
The increasing amount of data recorded by today's automatized and sensor-equipped production plants is an essential impetus for current developments in the industrial area. In order to face the challenge of actually making use of the recordings, machine learning algorithms can be employed to create models and hence, fully utilize the available data. In reference to a real-world production system, the relationships between multiple variables representing inputs (\cf configured process- and recipe parameters), internal states (\cf measured time series from condition monitoring) and dependent outputs (\cf measured product quality indicators) have to be covered by such models to identify a system comprehensively. One modeling approach for the analysis of complex systems are variable interaction networks \cite{rao2007variable} -- directed graphs representing system variables as nodes and their impact on others as weighted edges. Primarily, variable interaction networks have been employed to gain a better understanding of the interdependencies within a modeled system \cite{kommenda2011data}. In this work however, we utilize them to detect changing system behavior -- so-called concept drifts \cite{gama2014survey} -- online. Therefore, identified system relationships are tracked over time and analyzed for changes, which might give an indication for beginning malfunctions when applied at production plants.

The objective of this approach is closely related to the currently intensively investigated topic Predictive Maintenance \cite{lee2014service}, which is concerned with forecasting the remaining useful lifetime of a production system based on its current condition and scheduling specific preventing actions proactively. However, data which enables such predictions is quite difficult to gather, starting by consolidating data from various sources, up to carrying out a large number of run-to-failure experiments \cite{saxena2008damage} under continuous assessment of the actual system condition. Tracking of changing system relationships however, is applicable also for less stricktly controlled environments and allows a closer look into a system's dynamics.

In \Cref{sec:method}, we describe an algorithm to model variable interaction networks and present a sliding window based evaluation method, performing concept drift detection. Further on, a test problem is introduced in \Cref{sec:problem}, which we use to generate synthetic data streams and validate and discuss our approach in \Cref{sec:result}. Finally, we give a brief summary and an outlook for possible future extensions in \Cref{sec:conclution}.
%
%
%
\section{Variable Interaction Networks for Drift Detection}
\label{sec:method}
The developed two-phase concept drift detection approach may be categorized as supervised learning based detector \cite{krawczyk2017ensemble}. While the aim of the first phase is to develop a comprehensive model, which describes an initially stable system, during the second phase this network is used to detect structural interaction changes on a continuous stream of new data from the respective system. All parts of the approach have been implemented and tested using the open source framework HeuristicLab\footnote{\url{https://dev.heuristiclab.com/trac.fcgi/ticket/2288}}.
\subsection{Network Modeling}
\label{ssec:modeling}
As a first step, we define a set of variables within the system of interest. For each of them a regression model is trained using the other variables as inputs. For this task various machine learning algorithms may be employed, including multi-variate linear regression, random forests or symbolic regression. Subsequently, we determine the relevance of each input variable for the respective target within a model: The impact of a variable is calculated based on the increasing regression error of the developed model when re-evaluating it on a data set, for which the values of the variable have been randomly shuffled \cite{breiman2001random}. By this means, the information value of the respective variable is removed from the data set, without changing its distribution. Eventually, the calculated value is normalized to the range $0..1$. Subsequently, the directed, weighted graph for the variable interaction network can be constructed by creating a node for each participating variable and creating weighted edges from input to target nodes, by using the calculated impacts.

Several post-processing measures are advisable in order to prune less important nodes and edges and hence, determine more robust network structures: Regression models (and the derived variable impacts) with an estimation accuracy below a problem dependent threshold should not be taken into account right from the start, as they might not identify the system correctly. Further on, variable impacts below a user-defined threshold may be pruned to sparsify the networks, without loosing much information.

Moreover, we developed a routine, which uses the previously described impact computation as base, but assembles \emph{acyclic} graphs in order to support identifying the correct variable interaction \emph{direction}, as summarized in \autoref{lst:acyclic}. Instead of creating edges for any computed impact, the routine adds edges step-wise, alternating with removal of the weakest links to break up cycles.
\lstloadlanguages{csh}
\lstset{
	language=csh,
	basicstyle=\fontsize{8}{9}\ttfamily,
	numbers=left,
	numberstyle=\tiny,
	numbersep=10pt,
	tabsize=2,
	extendedchars=true,
	breaklines=true,
	frame=tb,
	captionpos=t,
	stringstyle=\color{blue}\ttfamily,
	showspaces=false,
	showtabs=false,
	xleftmargin=17pt,
	framexleftmargin=17pt,
	framexrightmargin=5pt,
	framexbottommargin=4pt,
	framextopmargin=4pt,
	commentstyle=\color{green},
	morecomment=[l]{//}, 
	morecomment=[s]{/*}{*/}, 
	showstringspaces=false,
	keywords={for, foreach, while},
	keywordstyle=\color{cyan},
	identifierstyle=\color{gray},
	backgroundcolor=\color{white},
	escapechar=?,
	columns=fullflexible
}
\begin{lstlisting}[label=lst:acyclic, caption={Creating an acyclic variable interaction network.}]
foreach variable // inputs and targets
?\indentrule?create a node

foreach node: create edge ?for? highest incoming impact
while any edges added // after loop: final acyclic graph
?\indentrule?find shortest cycles
?\indentrule?while any cycles found
?\indentrule?	?\indentrule?delete weakest link // based on impact
?\indentrule?	?\indentrule?find shortest cycles
?\indentrule?foreach node: create edge ?for? next highest incoming impact
\end{lstlisting}
%
\subsection{Network Evaluation}
\label{ssec:sweval}
\begin{figure}[h]
	\centering
	\includegraphics[width=.95\textwidth]{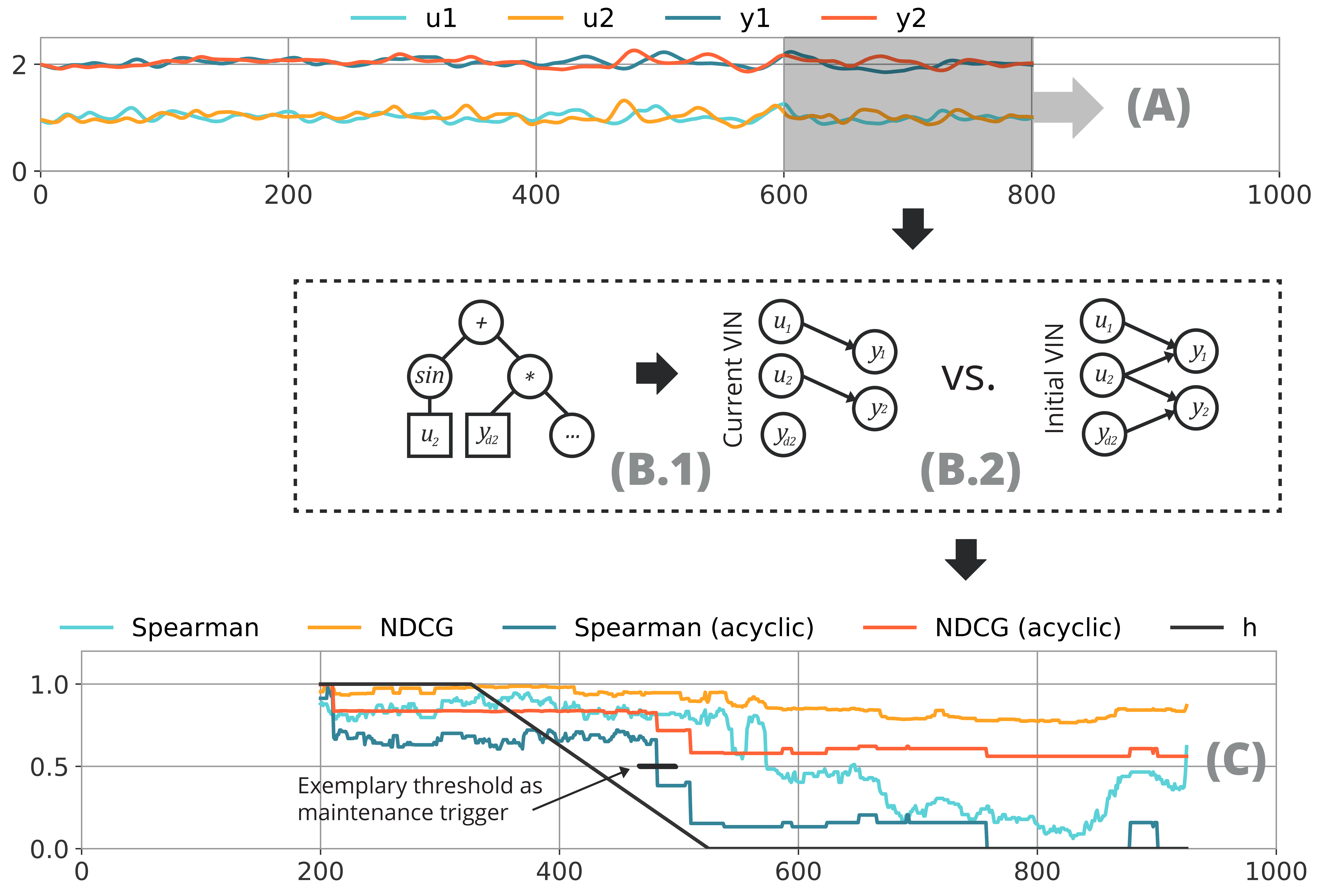}
	\caption{Sliding window based evaluation (A) of variable interaction networks (B.1, B.2) for concept drift detection (C). The decreasing network similarity scores and the also decreasing drift indicator $h$, represent the intended functioning of the proposed algorithm.}
	\label{fig:vineval}
\end{figure}
Within the second phase, as depicted in \autoref{fig:vineval}, we consider a stream of new, unseen data, which is partition-wise analyzed (A). The described calculation of variable impacts based on the previously built models, as well as the successive creation of networks is constantly repeated (B.1), while a window slides over this data stream. For identifying drifts in the underlying system, we compute the similarity of the initially built network -- representing a stable state -- and the updated networks, as part of the sliding window evaluation (B.2). Presumably, changing system behavior affects internal variable dependencies to some extent, which hence, should be reflected by the freshly created networks. We apply the Spearman’s rank correlation coefficient and the normalized discounted cumulative gain (NDCG), as proposed in \cite{kronberger2017measures}, to compare the network structures. The Spearman's rank correlation considers only deviations in ranks, such that top-ranked variables are treated equally to lower ranked variables. In contrast, the NDCG puts more weight on top-ranked variables by using an exponential weighting scheme.

A system may be declared \emph{drifting} if the similarity score during the evaluation drops below a threshold. If the actual drift state is known, as \eg for synthetic data sets, the correlation between the drift value and the computed similarity might give a good indication of how well the drift detection performed (C).
%
%
%
\section{Test Problem: Clogging Communicating Vessels}
\label{sec:problem}
\begin{figure}[h]
	\centering
	\includegraphics[width=.95\textwidth]{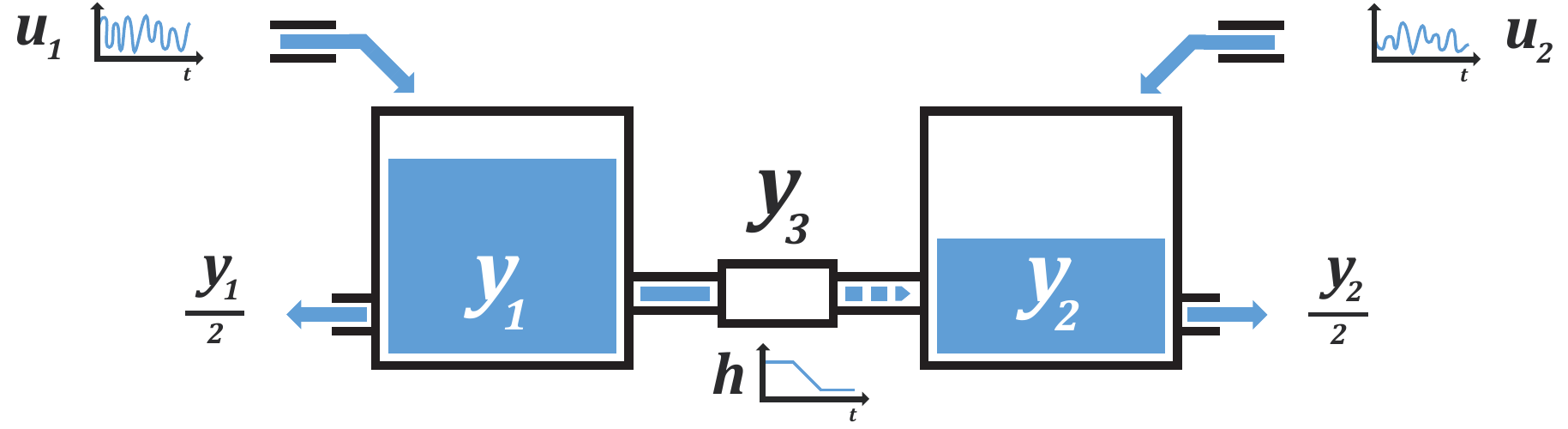}
	\caption{Vessel fill states $y_1$ and $y_2$ with inlets ($u_1$, $u_2$), outlets ($y_1/2$, $y_2/2$) and the gradually clogging (\cf $h$) communication channel with flow rate $y_3$.}
	\label{fig:commvess}
\end{figure}

In order to test the proposed concept drift detection algorithm, we designed a synthetic problem based on the system of communicating vessels as illustrated in \autoref{fig:commvess}. The vessels -- $y_1$ and $y_2$ represent their current fill state -- are continuously filled with fluid from two inlets. The flow rates of these inlets -- $u_1$ and $u_2$ -- are independent, dynamic and defined by stationary, auto-regressive models with normally distributed terms. The outlet flow rate for each vessel depends on the current fill state and hence, helps to preserve their stationarity. The communication channel between the vessels transports fluid into the vessel with the currently lower fill state and is described by the flow rate $y_3$. 
%
%
The dynamics of the system are defined by a system of differential equations (\ref{eq:y1}),(\ref{eq:y2}) and (\ref{eq:y3}). For this particular example we designed a channel that may gradually clog over time, thus, eventually resulting in a malfunctioning of the vessel communication, controlled by the parameter $h$ (\ref{eq:y3}). This clogging channel represents the maintenance problem, which is aimed to be found by the proposed detection algorithm.
\begin{align}
\label{eq:y1}
&\dot{y}_{1} = u_{1} + y_{3} - \frac{y_{1}}{2}&&&&\\[1mm]
\label{eq:y2}
&\dot{y}_{2} = u_{2} - y_{3} - \frac{y_{2}}{2}&&&&\\[1mm]
\label{eq:y3}
&\dot{y}_{3} = -(y_{1}-y_{2}) - hy_{3}&&&&
\end{align}
Based on the system definition we compiled a set of 10 training instances representing stable system states (\ie $h$ remains constant) and 10 evaluation instances with drifting behavior (\ie $h$ slowly decreases), each consisting of $1000$ data points. The variables allowed for model training and evaluation, as both inputs and targets, are $u_1$, $u_2$, $y_1$ and $y_2$. Further on, the first numerical derivative (as defined by the equations (\ref{eq:y1}) and (\ref{eq:y2})) and the second numerical derivative for each vessel fill state are provided as additional input variables. The current flow between the vessels, represented by $y_3$ and the clogging factor $h$ however, remain unknown to the regression models. This limitation is inspired by real-world problems, in which availability and quality of data are not always fully ensured, either for technical, monetary or security related reasons. It is the essential motivation and goal of the proposed drift detection algorithm to estimate and monitor the changing variable interactions, when this cannot be observed directly.
%
%
%
\section{Experiments and Results}
\label{sec:result}
The training of regression models, representing the variable interaction networks' foundation, was performed with multi-variate linear regression (LR), random forest (RF) and symbolic regression (SR). To tune the random forest and the symbolic regression algorithm, we performed a parameter grid search for reasonable configurations:
\begin{itemize}
	\item Random Forest: R: 0.5, M: 0.2, 100 trees
	\item Symbolic Regression: Offspring Selection Genetic Algorithm (OSGA) \cite{affenzeller2009modern}, population size: 100, generations: 1000, selection pressure: 100, proportional and random selection, mutation rate: 25\%, crossover rate: 100\%, unary functions ($\sin, \exp, \log$), binary functions $(+, -, \times, \div)$, max. tree length: 25 nodes
\end{itemize}
The modeling results, aggregated for all $10$ training instances, are summarized in \autoref{tab:qualities0}. Linear and symbolic regression both achieved almost perfect fits on the training as well as the test partition. The random forest models however, tend to overfit, no matter the tested algorithm parameters.
\vspace{-2mm}
\begin{table}[H]
	\centering
	\caption{Model estimation qualities $R^2$ and Normalized Mean Squared Error (NMSE) for the selected targets on training\,/\,test partition (\cf split at 66\%\,/\,34\%).}
	\label{tab:qualities0}
	\footnotesize \vspace{-1mm}
	\begin{tabular}{
			L{0.1\textwidth-2\tabcolsep}
			C{0.15\textwidth-2\tabcolsep}
			C{0.15\textwidth-2\tabcolsep-1.25\arrayrulewidth}|
			C{0.15\textwidth-2\tabcolsep}
			C{0.15\textwidth-2\tabcolsep-1.25\arrayrulewidth}|
			C{0.15\textwidth-2\tabcolsep}
			C{0.15\textwidth-2\tabcolsep}}
		\hline\noalign{\smallskip}
		& \multicolumn{2}{c}{Linear Regression (LR)} & \multicolumn{2}{c}{Random Forest (RF)} & \multicolumn{2}{c}{Symbolic Regression (SR)} \\
		\vspace{1mm}
		& $R^2$ & NMSE & $R^2$ & NMSE & $R^2$ & NMSE \\
		\noalign{\smallskip}\hline\noalign{\smallskip}
		$u_1$		& 0.99\,/\,0.99 & 0.00\,/\,0.00 & 0.96\,/\,0.76 & 0.06\,/\,0.30 & 0.97\,/\,0.97 & 0.02\,/\,0.03 \\
		$u_2$		& 0.99\,/\,0.99 & 0.00\,/\,0.00 & 0.95\,/\,0.70 & 0.06\,/\,0.34 & 0.97\,/\,0.97 & 0.02\,/\,0.03 \\
		$y_1$		& 0.99\,/\,0.99 & 0.00\,/\,0.00 & 0.95\,/\,0.67 & 0.06\,/\,0.41 & 0.93\,/\,0.92 & 0.06\,/\,0.09 \\
		$y_2$		& 0.99\,/\,0.99 & 0.00\,/\,0.00 & 0.95\,/\,0.71 & 0.06\,/\,0.36 & 0.94\,/\,0.92 & 0.05\,/\,0.09 \\	
		\noalign{\smallskip}\hline	
	\end{tabular}
\end{table}
\vspace{-2mm}
Based on the regression models, the initial variable interaction networks, representing stable system behavior, have been computed. Further on, we defined a threshold for the minimum NMSE of $0.2$ a model has to achieve to be considered for the network creation. For the random forest model, the threshold has been set higher, to an NMSE of $0.5$, because in the first modeling step we observed that the predictive quality of RF is lower compared to SR and LR. Furthermore, we set a minimum variable impact threshold of $0.1$ to prune less important edges from the final networks. After the modeling phase (\cf Section \ref{ssec:modeling}), one cyclic and one acyclic network version for each algorithm and each of the 10 training sets has been created.
\begin{figure}[h]
	\centering
	\includegraphics[width=.98\textwidth]{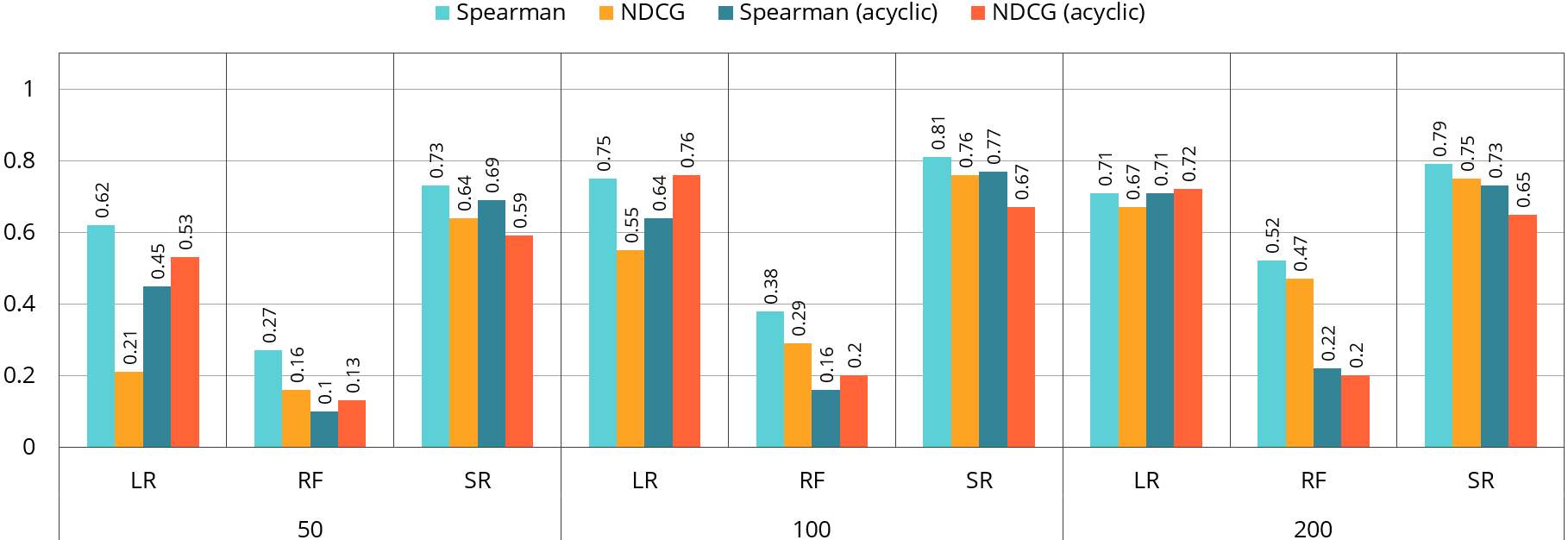}
	\caption{Correlation of the decreasing network similarity scores and the known, also decreasing drift indicator $h$, representing the gradually clogging communication channel.}
	\label{fig:qualitybarchart}
\end{figure}

The second phase (\cf Section \ref{ssec:sweval}) has been performed using the same configurations for the creation of networks during the sliding window evaluation. The results of the drift detection method for each regression modeling algorithm, with varying sliding window size and aggregated for all 10 drift data sets are depicted in \autoref{fig:qualitybarchart}. The bar chart illustrates the computed correlation between the network similarity and the synthetic drift, as described in \autoref{fig:vineval}.

According to the computed correlation scores, the linear and the symbolic regression models detect the synthetically introduced drifts quite well with a correlation of roughly $0.7$. Although the performance of the random forest models clearly lags behind, one can observe that the drifts are still detected to some extent. In conclusion, the detection algorithm is agnostic to the used regression models, however, accurate models with the ability to generalize (\ie not to overfit) are necessary.

One key factor of the detection algorithm is the sliding window size, which has to be tuned for any problem. With a large window size more stable network structures can be identified, which are valid for a longer period while moving over the data stream. This results in a smoother curve shape of the similarity score, however decreases the reaction speed to underlying trends and hence, should be limitted to a reasonable level. In this example, sizes between 100 and 200 showed similar good results. Furthermore, the acyclic networks achieved smoother similarity score curve shapes (\cf \autoref{fig:vineval}), than networks with cycles. Although, there is no advantage according to the computed detection quality by using these networks, it is easier to define a threshold as a minimum similarity score, when the values do not vary too much within a certain period, which is a clear benefit of the acyclic networks.
%
%
%
\section{Conclusion and Outlook}
\label{sec:conclution}
In this work we presented a machine learning based approach for identifying changing relationships of dynamical systems, such as industrial production plants. We show how variable interaction networks are developed and utilized to evaluate a continuous data stream and identify deviations from the original behavior, which eventually might enable triggering maintenance actions proactively. We implemented the algorighm using the open source framework HeuristicLab and tested the approach on a synthetic problem successfully. 

As a promising next step to enhance the described approach, we consider to investigate how a closer integration of modeling and evaluation phase might lead to a more accurate calculation of variable impacts and hence, more robust networks. A repeated or open-ended training of regression models -- which represent the foundation of the variable interaction networks -- on the continuously updated data stream might provide valuable information concerning the current impact of variables. Proceeding from drift detection, investigating the dependency changes closely, might eventually enable tracking a system change back to its beginnings. Especially considering the potential value for domain experts, such a root-cause analysis would be a powerful component for future production systems.
%
%
%
\subsubsection*{Acknowledgments}
The work described in this paper was done within the project ``Smart Factory Lab'' which is funded by the European Fund for Regional Development (EFRE) and the country of Upper Austria as part of the program ``Investing in Growth and Jobs 2014-2020''.
\vspace{-4mm}
\begin{figure}
	\centering
	\includegraphics[width=.70\textwidth]{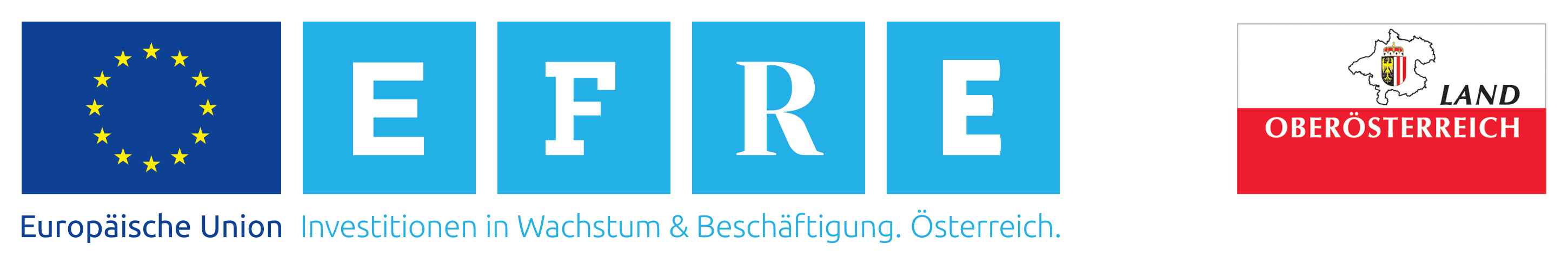}
	\label{fig:logo}
\end{figure}
\vspace{-4mm}

\noindent
Gabriel Kronberger gratefully acknowledges the financial support by the Austrian Federal Ministry for Digital and Economic Affairs and the National Foundation for Research, Technology and Development within the Josef Ressel Centre for Symbolic Regression.
\bibliographystyle{splncs04}
\bibliography{eurocast2019_zenisek}
\end{document}